\documentclass{article}

\usepackage{arxiv}
\usepackage[utf8]{inputenc} 
\usepackage[T1]{fontenc}    
\usepackage{hyperref}       
\usepackage{url}            
\usepackage{booktabs}       
\usepackage{amsfonts}       
\usepackage{nicefrac}       
\usepackage{microtype}      
\usepackage{subfigure} 
\usepackage{amsmath}        
\usepackage{lipsum}
\usepackage{graphicx}
\usepackage{algorithm}
\usepackage{algorithmic}

\title{Quantum Enhanced Multi-Scale CNN with Bi-Directional Mamba For Crop Field Analysis}

\author{
 Mohammad Salman Khan \\
  Department of Computer Science\\
  Lakehead University\\
  Thunder Bay, ON P7B5E1, Canada \\
   \And
 Ehsan Atoofian \\
  Department of Electrical and Computer Engineering\\
  Lakehead University\\
  Thunder Bay, ON P7B5E1, Canada \\
  \And
 Saad B. Ahmed* \\
  Department of Computer Science\\
  Lakehead University\\
  Thunder Bay, ON P7B5E1, Canada \\
 *Corresponding author \texttt{\{sbinahm@lakeheadu.ca\}} \\
}

\begin{document}
\maketitle
\begin{abstract}
Hyperspectral image (HSI) crop analysis plays a vital role in precision agriculture due to its ability to capture rich spectral and spatial information for accurate crop monitoring and assessment. However, effective HSI classification remains challenging because of high spectral dimensionality, spatial complexity, class imbalance, and limited labeled samples. To address these challenges, this work proposes a BiSpectral Mamba-based framework for HSI crop analysis that integrates multi-scale convolutional feature extraction, spectral attention, bidirectional state-space modeling, and quantum-inspired learning. The proposed architecture first employs a multi-scale CNN backbone to extract hierarchical spatial--spectral representations through feature fusion across multiple resolutions. A spectral channel attention mechanism is then utilized to emphasize informative spectral bands while suppressing redundant and noisy channels. The refined features are subsequently processed using a BiSpectral Mamba module, which models long-range dependencies in both directions by treating hyperspectral feature maps as sequential tokens. Furthermore, class-weighted optimization strategies and carefully designed fusion mechanisms are incorporated to improve training stability and classification performance under class imbalance conditions. Experimental results on the UAV-HSI-Crop dataset demonstrate that the proposed framework achieves an overall accuracy (OA) of $84.83\%$, validating the effectiveness of the architectural fusion for robust hyperspectral crop classification. The study further indicates that future performance improvements may be achieved by increasing the number of quantum qubits and integrating dynamic, class-aware spatial priors during the final bilinear upsampling stage. The proposed framework provides a promising foundation for advanced agricultural applications, including crop disease analysis, crop yield prediction, and soil moisture estimation, while also demonstrating the broader potential of structured state-space and quantum-inspired architectures for remote sensing and computer vision tasks.
\end{abstract}


\section{Introduction}
The rapid proliferation of Unmanned Aerial Vehicles (UAV) technology has transformed the field of remote sensing, allowing us to capture Hyperspectral Images (HSI) which have much more spectral data than the traditional RGB images. HSI provides a contiguous spectrum for every pixel, typically spanning hundreds of narrow spectral bands. These dense data cubes can be used for the precision identification of materials and biological states, as each emits unique spectral information, and helps us to better understand crop health, soil moisture and mineral composition.

Hyperspectral imaging (HSI) serves as a highly advanced tool for precision agriculture and remote sensing. This technology captures detailed spectral-spatial information across hundreds of contiguous wavelengths. Each material within an agricultural landscape exhibits a unique spectral signature. HSI encodes these fine-grained reflectance characteristics to reveal subtle physiological differences in vegetation that remain invisible to standard RGB sensors. The combination of high spectral resolution and precise spatial detail fundamentally transforms our ability to monitor food security, assess crop health, and optimize land use. High-resolution hyperspectral UAV imagery accurately captures both the spectral and spatial heterogeneity of these complex agricultural environments.
Analyzing these rich hyperspectral datasets introduces significant computational and mathematical challenges. UAV-based hyperspectral data inherently exhibit extreme high dimensionality, often containing up to $200$ distinct spectral bands. 
Agricultural datasets frequently suffer from a severe lack of labeled samples and display strong class imbalances.

Deep learning architectures provide powerful mechanisms for processing this complex spectral-spatial information. Convolutional neural networks extract hierarchical spatial features, whereas transformer encoder-decoder models successfully capture global dependencies across the image. State-of-the-art models achieve strong classification accuracy by combining spectral-attention modules with transformer blocks. These purely classical deep learning models require massive computational resources and carry exceptionally high parameter counts, sometimes approaching $100$ million parameters. Standard convolutional networks and transformers face significant difficulties in efficiently extracting meaningful structural information under the constraints of limited data and high dimensionality.

This work proposes the development and evaluation of highly efficient hybrid deep learning frameworks to overcome these computational barriers. 
The research explores the integration of multi-scale convolutional neural networks~\cite{multiscalecnn}, bidirectional Mamba state-space models, and variational quantum circuit~\cite{bidimamba}~\cite{quantumefficiency}. Multi-scale CNNs perform rapid, localized feature extraction, while bidirectional Mamba blocks enable efficient sequence-based modeling of the spectral data. Variational quantum circuits act as global feature enhancement heads to model non-linear global dependencies across the image patches. This hybrid quantum-classical approach significantly reduces the overall parameter count while maintaining highly competitive classification accuracy.

The primary objective is to validate these lightweight architectures in crop datasets. The training methodology employs a robust hybrid Cross-Entropy and Log-Cosh Dice loss function ~\cite{cnn6} to directly combat the extreme class imbalances inherent in the data. This specialized loss function optimizes the intersection-over-union for minority classes and stabilizes the gradient flow during training. Ultimately, this research establishes a scalable, computationally efficient paradigm for advanced crop field analysis in precision agriculture.

\section{Literature Review}

In this section, we will look at some of the recent research that has been done in deep learning applications for HSI images in crop analysis, and we will also look closely at how advanced neural network architectures like CNN, Mamba, and Quantum models work with HSI crop analysis datasets and what are their drawbacks.

\subsection{CNN for HSI Crop Analysis}

This section explains some of the recent works in CNN architecture for HSI crop analysis and identifies some of the gaps exhibited by CNN architecture in processing hyperspectral imagery.
Some benchmark HSI datasets for crop analysis are available, which feature a variety of objects in the scenes they capture. CMTNet~\cite{cnn1}, a CNN-Transformer hybrid, excels in UAV multi-crop classification and achieved $99.58\%$ overall accuracy on the WHU-Hi-LongKou dataset, using CNN branches for local spectral-spatial extraction fused with transformer global context---surpassing pure CNNs by $0.19--2.52\%$ across three UAV datasets. Ablin et al.~\cite{cnn2} proposed a lightweight ensemble architecture with additional quantum-inspired optimization. Their method reports strong performance on UAV-HSI-Crop of around $89\%$ OA. HRS-UNET~\cite{cnn5} uses a U-Net style encoder--decoder with convolutional blocks that also combines Multiscale Spectral Aggregation (MSA) module to compress spectral dimension while enhancing features and reducing computation, reporting OA of $89.96\%$ and kappa score of $0.8814$.There are also reported some comparison scores on basic CNN models like SegNet and UNet~\cite{cnn6} which reported accuracies of $43.61\%$ and $76.07\%$ respectively. 

A 3D-CNN for for soybean disease~\cite{cnn7} used Pure 3D CNN taking full HSI cube (spatial + spectral) as input and several 3D conv + pooling layers, followed by FC classification for ``healthy vs diseased'' soybean (charcoal rot) on a dataset of Hypsectral cubes of soybean stems and reported accruacies of OA $95.73\%$ and infected class F-1 score of $0.87$.In MLVI-CNN~\cite{cnn9} HSI is preprocessed using Savitzky--Golay, normalization and band selection is done via RFE to derive MLVI and H-VSI indices which are processed using 1D CNN on index sequences on a dataset of HSI Crop stress experiment reporting accuracies of $83.40 \%$ OA.

\subsection{Transformers for HSI Crop Analysis}

This section compiles recent work on the Transformer architecture for HSI crop analysis and identifies some gaps in the architecture.
Vision transformers and HSI specialized variants of them, have proved to be good for crop analysis as they can effectively capture spectral-spatial dependencies through self-attention, addressing CNNs fixed receptive field limitations. Recent works demonstrate competitive accuracies on agricultural benchmarks while capturing long-range crop-field relationships.

Niu et al.~\cite{cnn6} introduced HSI-TransUNet, a transformer-based encoder-decoder for crop mapping on UAV-HSI-Crop, reporting OA of $86.05\%$ and kappa of $0.8347$. Hossain et al.~\cite{trans2} studied transformer-UNet adaptation for UAV agricultural HSI and reported OA of $76.6\%$ with kappa of $0.532$.The BandWiseTransformer~\cite{trans10} architecture is a one-dimensional self-attention network featuring an embedding layer, four transformer encoder blocks with multi-head attention, and an adaptive average pooling layer tailored to capture fine-grained spectral dependencies.It was evaluated on a custom UAV hyperspectral library of $3,105$ plot-level wheat spectra representing five different diseases, achieving a reported overall test accuracy of $97.9\%$. 

The Semantic-Guided Transformer Network (SGTN)~\cite{trans3} introduces semantic guidance to reduce background interference and emphasize crop regions in HSI, achieving OA $98\%$ on Indian Pines, Pavia University, and Salinas. HyperSFormer~\cite{trans4} uses an end-to-end transformer that treats HSI pixels as tokens, with multi-head self-attention over spectral--spatial features and an adaptive min-log sampling (AMLS) strategy and Dice+focal loss to handle insufficient and imbalanced samples and reports accuracies of $98.4\%$ OA on the Indian Pines Dataset. MSA-LWFormer~\cite{trans5} uses multi-scale spectral attention and FFT-based frequency domain Transformer that reduces complexity and reported OAs of $98.87\%$ , $99.79\%$ and $99.96\%$ on Indian Pines , Pavia University and Salinas datasets respectively. SSSAT-Net~\cite{trasn6} integrates PCA for dimensionality reduction, a Convolutional Block Attention Module, Spectral and Spatial attention modules, and a Transformer to comprehensively extract local and global spectral-spatial features.It was evaluated on the Indian Pines, Kennedy Space Center, Pavia University, Houston 2013, and Salinas datasets,achieving overall accuracies of $98.75\%$, $99.42\%$, $99.92\%$, $99.42\%$, and $99.91\%$, respectively. A locally enhanced transformer network featuring a multibranch spatial--spectral tokenization module ~\cite{trans7} for shallow feature extraction and a dual-branch transformer encoder to capture both local and global spatial dependencies.It was evaluated on the Pavia University, University of Houston, WHU-Hi-LongKou, and ShanDongFeiCheng datasets, achieving reported overall accuracies of $96.66\%$, $90.12\%$, $98.24\%$, and $61.29\%$,respectively. The Efficient Attention Transformer Network (EATN)~\cite{trans8} utilizes a Self-Similarity Feature Enhancement (SSFE) module combined with Spectral Interactive Transformer (SIT) and Spatial Conv-Attention (SCA) modules to efficiently extract local and global spatial-spectral features. It was evaluated on the Houston 2013, achieving a reported overall accuracy of $95.96\%$ on the Houston 2013 dataset. The Low-rank Transformer Network (LRTN)~\cite{trans9} utilizes a U-shaped architecture featuring dual encoding branches with spectral cross attention and spatial low rank cross attention, combined with a spatial-spectral priors guided fusion module achieving $78.1\%$ OA on Houston Dataset.

~~~~~While a transformer has many advantages, it relies heavily on global self-attention, and that has quadratic complexity based on the number of tokens. For HSI data that is high in dimensions, this presents issues related to computational efficiency and larger memory requirements. Additionally, Transformers treats inputs as sequences of tokens and may not preserve the inherent spatial topology of agricultural scenes.

\subsection{Quantum Methods for HSI Crop Analysis}

This section presents the recent works in quantum-based architecture for HSI crop analysis and identifies some of the gaps in them.
Quantum models possess a fundamental architectural advantage for processing highly complex agricultural data. Hyperspectral imagery generates massively high-dimensional datasets requiring extensive computational power to map global dependencies across an entire farm field. Variational quantum circuits natively model these complex feature interactions through the quantum mechanical principles of superposition and entanglement. This specialized quantum mechanics enables the highly efficient representation of non-linear global dependencies across all spatial patches of the hyperspectral image. The quantum circuit forces the neural network to evaluate holistic relationships simultaneously. This simultaneous processing capability drastically improves the model's ability to recognize the incredibly subtle spectral differences between highly similar crop classes

A hybrid model architecture using a MobileNetV2 backbone for classical feature extraction with a quantum-inspired processing layer that simulates variational quantum circuits was evaluated on a custom dataset of approximately $5,000$ labeled fruit crop leaf images (categorized as healthy, wilt-affected, and nematode-infested), the hybrid model achieved a reported overall accuracy of $98.8\%$~\cite{quant1}. A model ~\cite{quant2} which utilizes a D-Wave quantum annealer for Mutual Information (MI)-based spectral band selection ( followed by quantum boosting algorithms (Qboost and Qboost-Plus) that use Error-Correcting Output Codes for multi-label classification was evaluated on the AVIRIS Indian Pine hyperspectral dataset, the quantum classifiers reportedly achieved high accuracies that outperformed conventional baselines such as Adaboost and SVM, although specific overall accuracy percentages are not provided. An architecture that resolves the multispectral unmixing problem by first generating a virtual hyperspectral image, and then performing unmixing utilizing a Weighted Simplex Shrinkage (WSS) regularizer alongside a Quantum Deep Image Prior (QDIP) module was evaluated on AVIRIS (Boulder, Ottawa, Hawaii) and Sentinel-2 remote sensing datasets, the algorithm unmixing performance with reported average error of 8.193 for RMSE~\cite{quant3}. The QSSPN~\cite{quant4} architecture cascades multiple QSSN blocks---each containing a phase-prediction module and a measurement-like fusion module---to dynamically extract and jointly fuse spectral-spatial features. It was evaluated on the Indian Pines, Pavia University, and Salinas hyperspectral datasets with overall accuracies of $95.87\%$, $99.71\%$, and $99.66\%$, respectively. Liu et al.~\cite{quant5} explored quantum-assisted HSI classification with a quantum annealing SVM on limited-label datasets. Lin and Young ~\cite{quant6} introduced a hybrid quantum-classical GAN for hyperspectral restoration. 
Overall, quantum methods in remote sensing are promising but remain underexplored for UAV crop-field classification.
  
~~~~~Current quantum machine learning research exhibits critical limitations regarding hyperspectral agricultural classification. Existing studies primarily apply quantum algorithms to scarcely labeled generic remote sensing datasets or large-scale satellite imagery rather than specialized agricultural data. The exploration of hybrid quantum models for highly imbalanced, real-world crop field datasets has not been addressed. No research has been done on combining advanced spatial modeling techniques such as graph-attention networks with these quantum implementations within agricultural contexts.

\subsection{Mamba and State-Space Models for HSI Crop Analysis}
This section explains the Mamba-based architecture for HSI crop analysis.
Mamba state-space models deliver incredibly fast sequence processing for high-dimensional hyperspectral data. The architecture structurally treats the extracted multi-scale spatial features and continuous spectral bands as an extended, flattened data sequence. This advanced sequence-based feature extraction provides deep contextual understanding without triggering the massive quadratic memory bottlenecks inherent to standard Transformers. The model scales linearly with sequence length, establishing it as a highly efficient and structurally superior engine for parsing dense, multi-band agricultural environments.

The SS-Mamba~\cite{mamba1} architecture consists of a token generation module and stacked dual-branch Mamba blocks equipped with a feature enhancement module to efficiently extract and fuse long-range spatial and spectral sequences. It was evaluated on the Indian Pines, Pavia University, Houston, and Chikusei hyperspectral datasets, achieving reported overall accuracies of $91.59\%$, $96.40\%$, $94.30\%$, and $94.97\%$,respectively. The HyperspectralMamba~\cite{mamba2} architecture features a dual stream design that combines a State Space Model(SSM) to capture global spectral dependencies with a parallel 1D convolutional path to extract local features, which are then integrated using a band-adaptive feature recalibration mechanism. It was evaluated on the Indian Pines, Pavia University, and Salinas Valley datasets, and it achieved reported overall accuracies of $95.31\%$, $98.60\%$, and $96.40\%$, respectively. The BiMambaHSI architecture is a bidirectional state-space framework that integrates a joint spectral-spatial gated Mamba(JGM) encoder and a dual-branch spatial-spectral Mamba block(SSMB) to explicitly capture forward-backward long-range dependencies while maintaining linear complexity.It was evaluated on the Pavia University, Houston, Indian Pines, WHU-Hi-HanChuan, and WHU-Hi-LongKou hyperspectral datasets, achieving reported overall accuracies of $97.90\%$, $97.92\%$, $99.02\%$, $97.92\%$, and $99.54\%$,respectively. The EchoMamba~\cite{mamba4} combines a Random Forest SMOTE(RFMS) data preprocessing strategy with a novel LSTMS6 network, which cascades an LSTM layer to extract local temporal dependencies and a Mamba(S6) module to encode global spectral context. It was evaluated on the Augsburg, Salinas, Pavia Centre scene, Pavia University, Indian Pines, and Houston 2013 hyperspectral datasets; it achieved reported overall test accuracies of $99.70\%$, $98.65\%$, $95.44\%$, $99.72\%$, $99.03\%$, and $98.40\%$, respectively.
MorpMamba integrates~\cite{mamba5} morphological operations to generate spatial-spectral tokens, refines them using multihead self-attention, and processes them through a linear-complexity State Space Model(SSM) for efficient hyperspectral image classification. It was evaluated on the WHU-Hi-LongKou, Pavia University, Pavia Centre, Salinas, and University of Houston datasets, its optimal spatial-spectral variant (SSMM) achieved reported overall accuracies of $99.70\%$, $97.67\%$, $99.71\%$, $98.52\%$, and $98.28\%$, respectively.

The S2Mamba~\cite{mamba6} uses two linear-complexity selective structured state space models, a Patch Cross Scanning module for spatial features and a Bi-directional Spectral Scanning module for spectral features dynamically fused via a Spatial-spectral Mixture Gate. It was evaluated on the Indian Pines, Pavia University, and Houston 2013 datasets, it achieved reported overall accuracies of $97.92\%$, $97.81\%$, and $93.36\%$,respectively. The MambaMoE~\cite{mamba7} is an encoder-decoder framework featuring a Mixture of Mamba Expert Blocks, which uses spatial routed and spectral shared experts combined with an uncertainty-guided corrective learning strategy to dynamically extract joint spectral-spatial features. It was evaluated on the Pavia University, Houston, and Whu-HanChuan hyperspectral datasets, and it achieved reported overall accuracies of $95.20\%$, $91.18\%$, and $92.67\%$, respectively. The SSUMamba ~\cite{mamba8} is an encoder-decoder framework designed for hyperspectral image denoising that utilizes Spatial-Spectral Continuous Scan (SSCS) Mamba blocks combining bidirectional State Space Models with 3D residual convolutions to efficiently capture both local textures and long-range spatial-spectral dependencies. It was evaluated on the ICVL, Houston 2018, Pavia City Center, Gaofen-5 Wuhan, and Earth Observing-1 datasets. The model is measured on denoising performance rather than classification accuracy, achieving noise ratios (PSNR) of $43.07$, $34.74$, and $35.70$ under mixture noise conditions on the synthetic datasets, respectively.

~~~~~Mamba models fundamentally struggle to preserve the native spatial geometry of agricultural fields. State-space architectures require flattening the three-dimensional hyperspectral data cube into a single, one-dimensional sequence for processing. This aggressive unrolling process actively destroys the critical local boundaries between distinct crop rows and adjacent diseased zones. Mamba systems aggressively prioritize tracking spectral signatures across sequential bands at the direct expense of spatial awareness. Pure sequential state-space modeling forces the network to lose vital localized context and easily misclassifies small, isolated pathogen outbreaks

\subsection{Gaps Identified}

A comprehensive review of the current literature reveals specific, critical failures across every major deep learning architecture when applied to hyperspectral crop analysis.

\begin{itemize}
   
\item \textit{Convolutional Neural Networks (CNNs)}:
Standard convolutional models rapidly exhaust computational limits when scaled to process high-volume hyperspectral data. Lightweight CNN variants completely fail to capture the holistic, global structural dependencies required to differentiate spectrally similar crop species. Traditional local receptive fields consistently struggle to maintain precise boundary delineations between dense, overlapping vegetation rows.

\item \textit{Transformers}:
Vision transformers introduce absolutely unsustainable parameter overheads, such as the 99.33 million parameters required by the HSI-TransUNet. These heavy classical architectures demand massive volumes of explicitly labeled data to converge properly, a resource entirely absent in real-world agricultural environments. 
Deep transformer networks consistently over-smooth localized, high-frequency spatial textures crucial for identifying small, isolated crop clusters

\item \textit{Quantum Based Model}:
Current quantum machine learning applications strictly analyze low-resolution satellite imagery or simplified, generic benchmark datasets
Existing studies completely fail to fuse quantum global feature extractors with advanced localized spatial modeling techniques.

\item \textit{Mamba Model}:
Standard state-space models aggressively flatten three-dimensional hyperspectral cubes into rigid one-dimensional sequences. This forced unrolling mechanism mathematically destroys the native spatial geometry and vital structural boundaries of the agricultural field. Pure sequential modeling completely sacrifices critical local neighborhood context to prioritize tracking continuous spectral signatures.
  
\end{itemize}

This work systematically addresses these precise architectural bottlenecks to deliver a highly functional and field-ready classification system. This work presents the CNN-BiSpectralMamba-Quantum model to completely bypass standard parameter bloat. Integrating a 4-qubit variational quantum circuit explicitly replaces massive classical fully-connected layers. This specific structural swap drops the total trainable parameter count to a remarkably lightweight.  
This work brought quantum machine learning directly into the precision agriculture domain by validating these hybrid structures on the complex, 30-class UAV-HSI-Crop dataset. To eliminate the Mamba architecture's spatial destruction, the work fused the bidirectional state-space blocks directly atop a multi-scale CNN feature extractor. The multi-scale convolutions strictly anchor the spatial geometry before the Mamba blocks process the complex spectral sequences. 

~~~~~Finally, the work uses an optimization pipeline to survive the dataset's extreme class imbalances. by implementing a hybrid Cross-Entropy and Log-Cosh Dice loss function to force the network to stabilize gradients and aggressively optimize the intersection-over-union metric for rare, minority crop classes.

\section{Dataset Overview}

This section provides a detailed analysis of the dataset utilized in this study. The UAV-HSI-Crop dataset serves as the primary foundation of the proposed work. This large-scale hyperspectral dataset was developed by researchers at China Agricultural University~\cite{cnn6} and contains rich spectral and spatial information collected from real agricultural environments, making it well-suited for evaluating advanced hyperspectral image classification frameworks. 
They flew specialized drones directly over dense agricultural fields located in Majiakou Village and Xijingmeng Village within Shenzhou City, Hebei Province, China. The collection team mounted a high-performance Resonon Pika-L hyperspectral sensor on the UAVs to capture the imagery. This sensor actively recorded spectral reflectance information across the precise $400$ to $1000$ nm wavelength range. Consequently, the hardware generated a massive, high-dimensional data cube containing $200$ contiguous spectral bands for every single flight. The resulting imagery boasts a razor-sharp spatial resolution of approximately $0.1$ meters per pixel. The detail is provided in Table~\ref{tab:dataset_summary}, which perfectly captures the complex structural variations and spatial heterogeneity of the actual crop fields.

\begin{table}[htb!]
    \centering
    \caption{Summary of the UAV-HSI-Crop dataset used in this work.}
    \footnotesize
    \label{tab:dataset_summary}
    \begin{tabular}{lc}
    \toprule
    Property & Value \\
    \midrule
    Samples & $433$ \\
    Spatial size per sample & $96 \times 96$ \\
    Spectral bands & $200$ \\
    Sensor wavelength range & $400--1000 nm$ \\
    Spatial resolution & $\sim 0.1$ m/pixel \\
    Number of classes & $27$ \\
    \bottomrule
    \end{tabular}
\end{table}

\begin{figure}[htb!]
    \centering
    \includegraphics[width=0.5\linewidth]{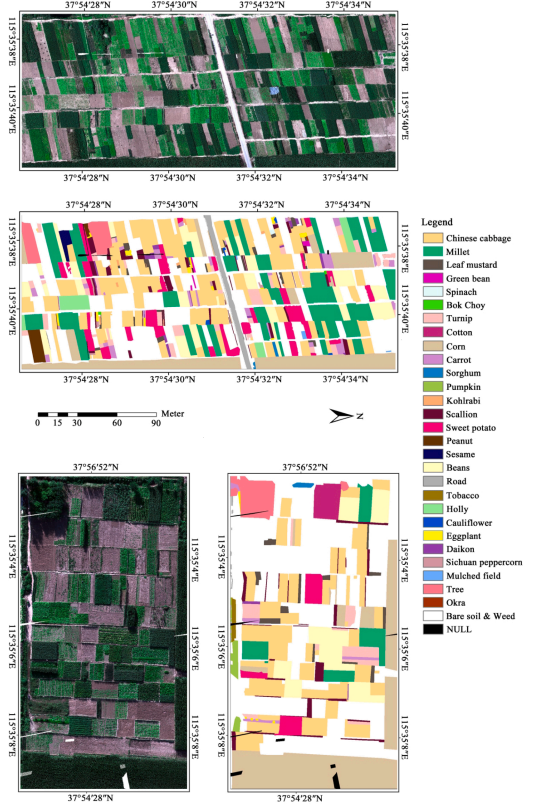}
    \caption{Classes in UAV-HSI-Crop Dataset as shown in ~\cite{cnn6}}
    \label{fig:datasetclasses}
\end{figure}

\begin{figure}[htb!]
    \centering
    \includegraphics[width=0.7\linewidth]{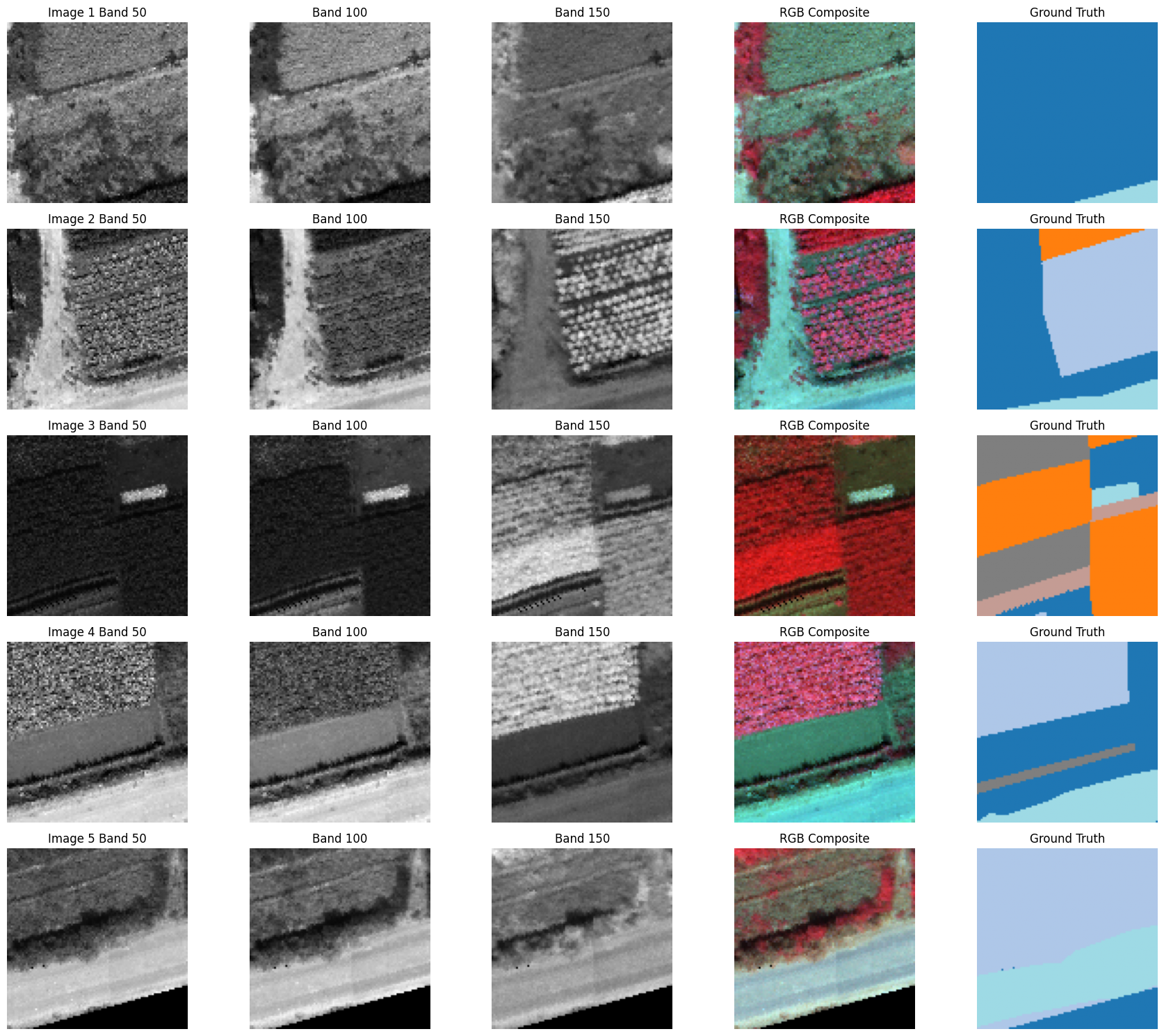}
    \caption{Examples from UAV-HSI-Crop: selected spectral bands, RGB composite, and ground-truth labels.}
    \label{fig:dataset_examples}
\end{figure}

This dataset provides an incredibly rigorous testing ground for deep learning architectures. The complete repository delivers 433 distinct hyperspectral image samples. Each sample consists of a highly detailed $96 \times 96$ pixel spatial grid packed with all $200$ spectral channels. The dataset meticulously categorizes 30 distinct crop classes as seen in Fig~\ref{fig:datasetclasses} and land cover types. The ground truth labels explicitly identify complex, overlapping categories including bare soil, invasive weeds, Chinese cabbage, corn, millet, and a wide variety of other specific vegetation classes as seen in Fig~\ref{fig:dataset_examples}. 
Real-world agricultural data inherently contains massive structural biases, and this dataset accurately reflects that reality through a brutally severe class imbalance. Dominant staple crops flood the pixel counts, while rare vegetation types barely register in the localized ground truth maps. This exact combination of extreme high-dimensionality, rich spatial resolution, and severe class imbalance makes the UAV-HSI-Crop dataset a good base to test our models on agricultural applications.

\section{Methodology}

\begin{figure}[htb!]
    \centering
    \includegraphics[width=1\linewidth]{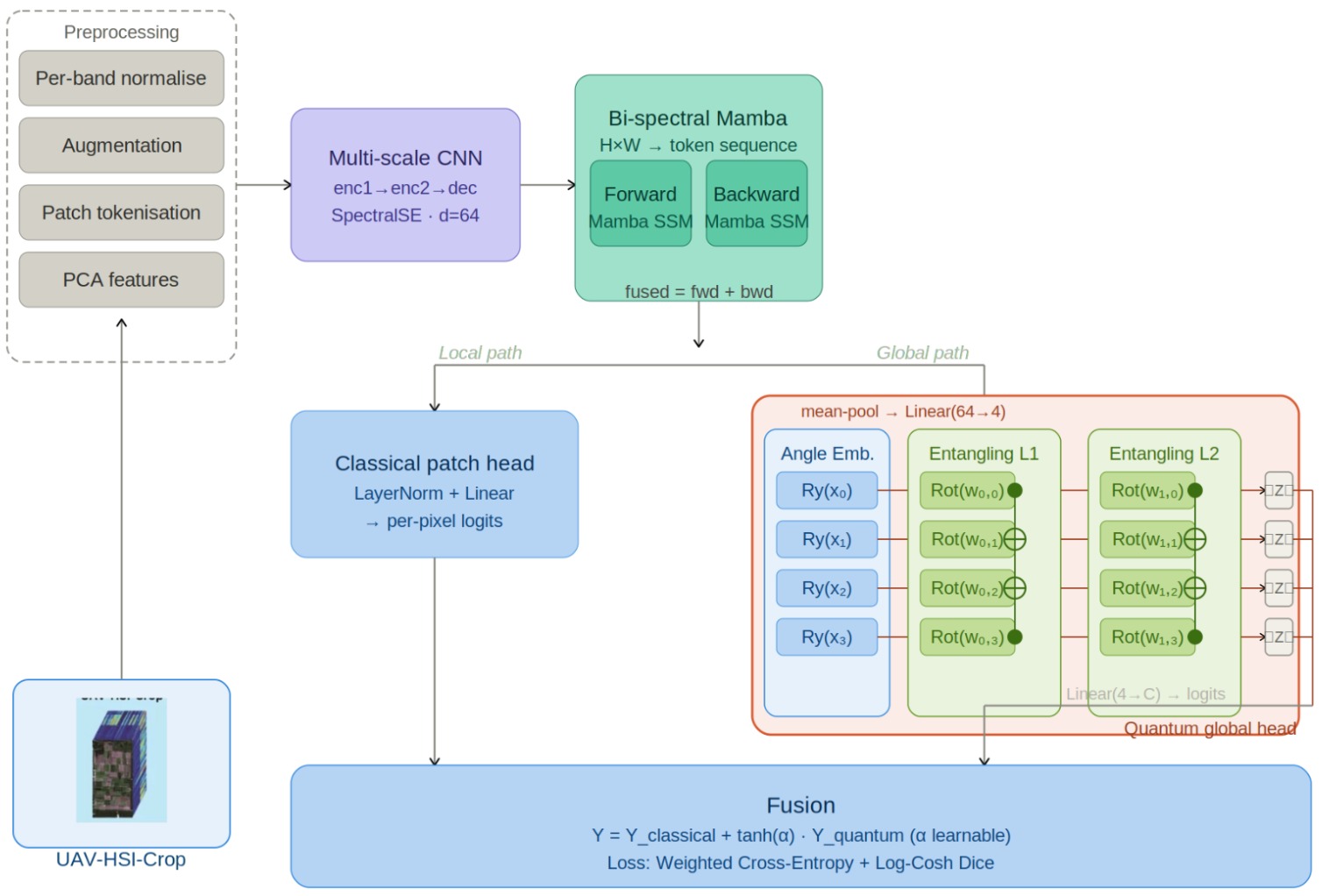}
    \caption{Methodology of Quantum Enhanced Multi-Scale CNN with Bi-Directional Mamba}
    \label{fig:work2method}
\end{figure}

In this section we will look at the architecture of the Quantum Enhanced Multi-Scale CNN with BiDirectional Mamba as show in Fig~\ref{fig:work2method} and the mathematical formulations behind its working.

The CNN-BiSpectralMamba-Quantum architecture establishes a high-speed, parameter-efficient pipeline for hyperspectral crop classification. This framework actively fuses multi-scale spatial convolutions, bidirectional state-space sequence modeling, and a variational quantum circuit.

\subsection{Data Normalization Pipeline}
The data pipeline aggressively standardizes the raw hyperspectral tensor $X \in \mathbb{R}^{B \times C \times H \times W}$ prior to any deep learning operations . The algorithm computes the precise mean $\mu_c$ and standard deviation $\sigma_c$ across the spatial dimensions for every single spectral band $c$ . The mathematical standardization strictly enforces a minimum standard deviation floor of $10^{-6}$ to guarantee absolute numerical stability.
\begin{equation}
    X_{norm}^{(c)} = \frac{X^{(c)} - \mu^{(c)}}{\max(\sigma^{(c)}, 10^{-6})}
\end{equation}

Rest of the preprocessing is similar the QPGF architecture and follow the same guidelines.

\begin{figure*}
    \centering
    \includegraphics[width=1.0\linewidth]{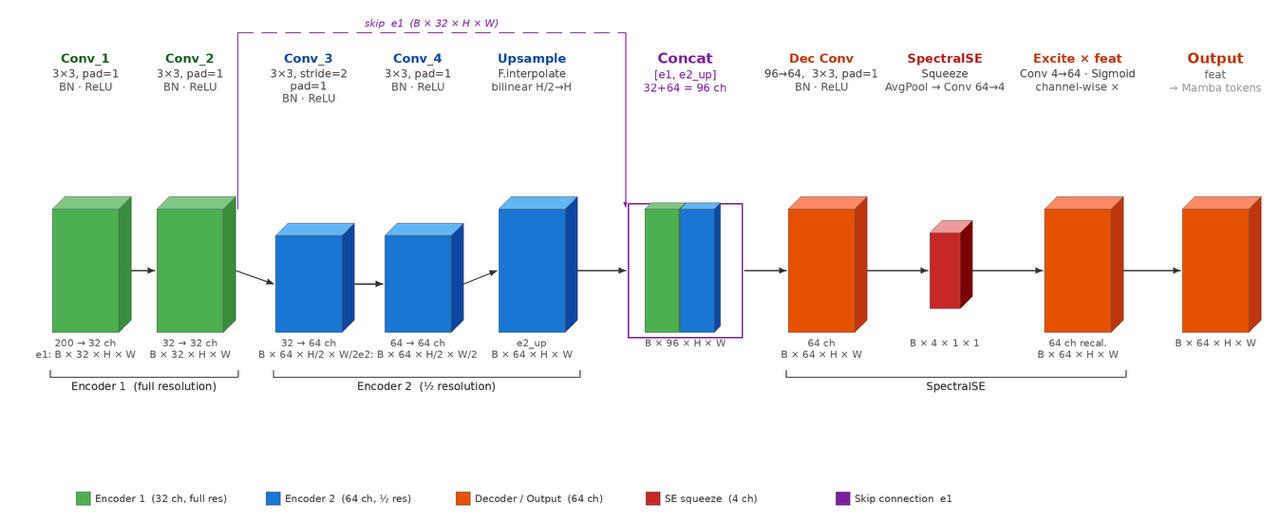}
    \caption{Multi-Scale CNN}
    \label{fig:mscnnarch}
\end{figure*}

\subsection{Vanilla Mamba}
\begin{figure}
    \centering
    \includegraphics[width=0.7\linewidth]{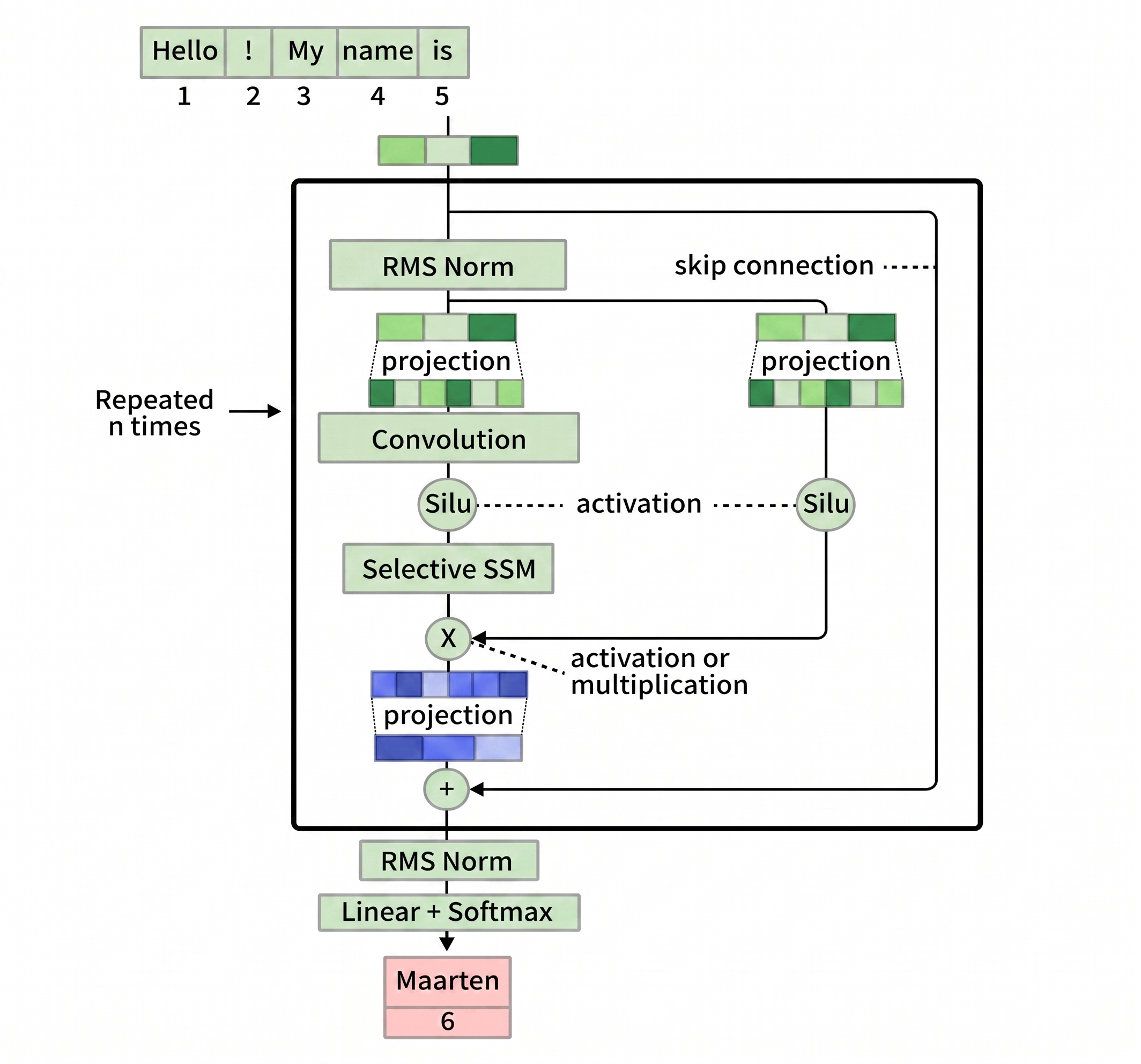}
    \caption{Vanilla Mamba~\cite{vanillamambba}}
    \label{fig:Vanilla Mamba}
\end{figure}

The Mamba model (as shown in Fig~\ref{fig:Vanilla Mamba}) was introduced as a model to tackle text problems in deep learning and it works by breaking down text sequences into individual tokens. The model transforms these discrete words into dense vector representations~\cite{vanillamambba}.

An embedding layer converts these IDs into an initial input matrix $X$. Let $x_t$ represent the vector for the token at position $t$.

\textit{Root Mean Square Normalization} stabilizes the learning process by scaling the features before they enter the main branches.
\begin{equation}
    u_{\text{norm}} = \frac{u}{\sqrt{\frac{1}{d} \sum_{i=1}^{d} u_i^2 + \epsilon}} \odot \gamma
\end{equation}

This normalized data splits into two parallel pathways. Left path, which handles local context and sequence modeling, and the right path, which acts as a gate to remember important context

In the left path, the projection, which is a linear transformation, expands the feature dimension of the input, which is denoted by $z_1 = W_1 u_{\text{norm}}$, and this is then followed by a 1D causal convolution that mixes local sequential information across adjacent tokens, creating a convolved sequence $c$.
The Sigmoid Linear Unit introduces non-linearity.

\begin{equation}
    \text{SiLU}(x) = x \cdot \sigma(x) = \frac{x}{1 + e^{-x}}
\end{equation}

The \textit{Selective State Space Model} acts as the core reasoning engine. It dynamically filters information based on the input context. The parameters $B$, $C$, and the step size $\Delta$ are derived from the input data, making the model selective. The continuous equations are discretized to process the sequence step-by-step
\begin{equation}
h_t = \bar{A} h_{t-1} + \bar{B} x_t
\end{equation}

\begin{equation}
 y_t = C h_t
\end{equation}

where $\bar{A}$ and $\bar{B}$ are discretized versions of the continuous state parameters, and $h_t$ is the hidden state.Let the output of this layer be $y$.

In the right path, the data passes through another SiLU function to create a gating signal. Let this be 
\begin{equation}
g = \text{SiLU}(z_2)     
\end{equation}

Element-wise multiplication merges the two branches. The right branch acts as a multiplicative gate, regulating the signals from the Selective SSM.
\begin{equation}
m = y \odot g  
\end{equation}

A linear layer projects the gated matrix back to the original block dimension. Let this be 
\begin{equation}
o = W_3 m    
\end{equation}

Addition (+): A residual connection adds the original block input $u$ to the final projection, allowing gradients to flow easily during training.
\begin{equation}
u_{\text{next}} = u + o
\end{equation}

This output can be passed to output heads such as softmax or a custom output head to get the predictions.

\subsection{Multi-Scale Convolutional Feature Extraction}
The normalized tensor passes directly into a Multi-Scale Convolutional Neural Network as shown in Fig~\ref{fig:mscnnarch}. The initial encoder block extracts foundational spatial textures using two sequential $3 \times 3$ convolutions, coupled with Batch Normalization and ReLU activations, to produce a 32-channel feature map $E_1$.
\begin{equation}
    E_1 = \text{ReLU}(\text{BN}(\text{Conv}_{3\times3}(\text{ReLU}(\text{BN}(\text{Conv}_{3\times3}(X_{norm}))))))
\end{equation}

The second encoder block explicitly halves the spatial resolution via a strided convolution (stride 2) and expands the feature depth to $d_{model} = 64$ to generate $E_2$ .
\begin{equation}
    E_2 = \text{ReLU}(\text{BN}(\text{Conv}_{3\times3}(\text{ReLU}(\text{BN}(\text{Conv}_{3\times3, s=2}(E_1))))))
\end{equation}

The decoder pathway formally upsamples $E_2$ using deterministic bilinear interpolation to match the spatial dimensions of $E_1$. The network securely concatenates these multi-scale tensors along the channel dimension and applies a final convolution to yield the fused spatial map $F_{dec}$.
\begin{equation}
    F_{dec} = \text{ReLU}(\text{BN}(\text{Conv}_{3\times3}([E_1 \parallel \text{BilinearUp}(E_2)])))
\end{equation}

\subsection{Spectral Squeeze-and-Excitation}

A Spectral Squeeze-and-Excitation block dynamically recalibrates the channel-wise feature responses. An Adaptive Average Pooling operation compresses the spatial grid into a $1 \times 1$ global descriptor. A two-layer fully connected network computes the excitation weights utilizing a reduction ratio of 16 . A Sigmoid activation function $\sigma_{sig}$ scales the input tensor $F_{dec}$ through direct element-wise multiplication.

\begin{equation}
    F_{SE} = F_{dec} \otimes \sigma_{sig}(W_2 \cdot \text{ReLU}(W_1 \cdot \text{AdaptiveAvgPool}(F_{dec})))
\end{equation}

\subsection{Bidirectional Mamba State-Space Mechanics}
\begin{figure}
    \centering
    \includegraphics[width=0.6\linewidth]{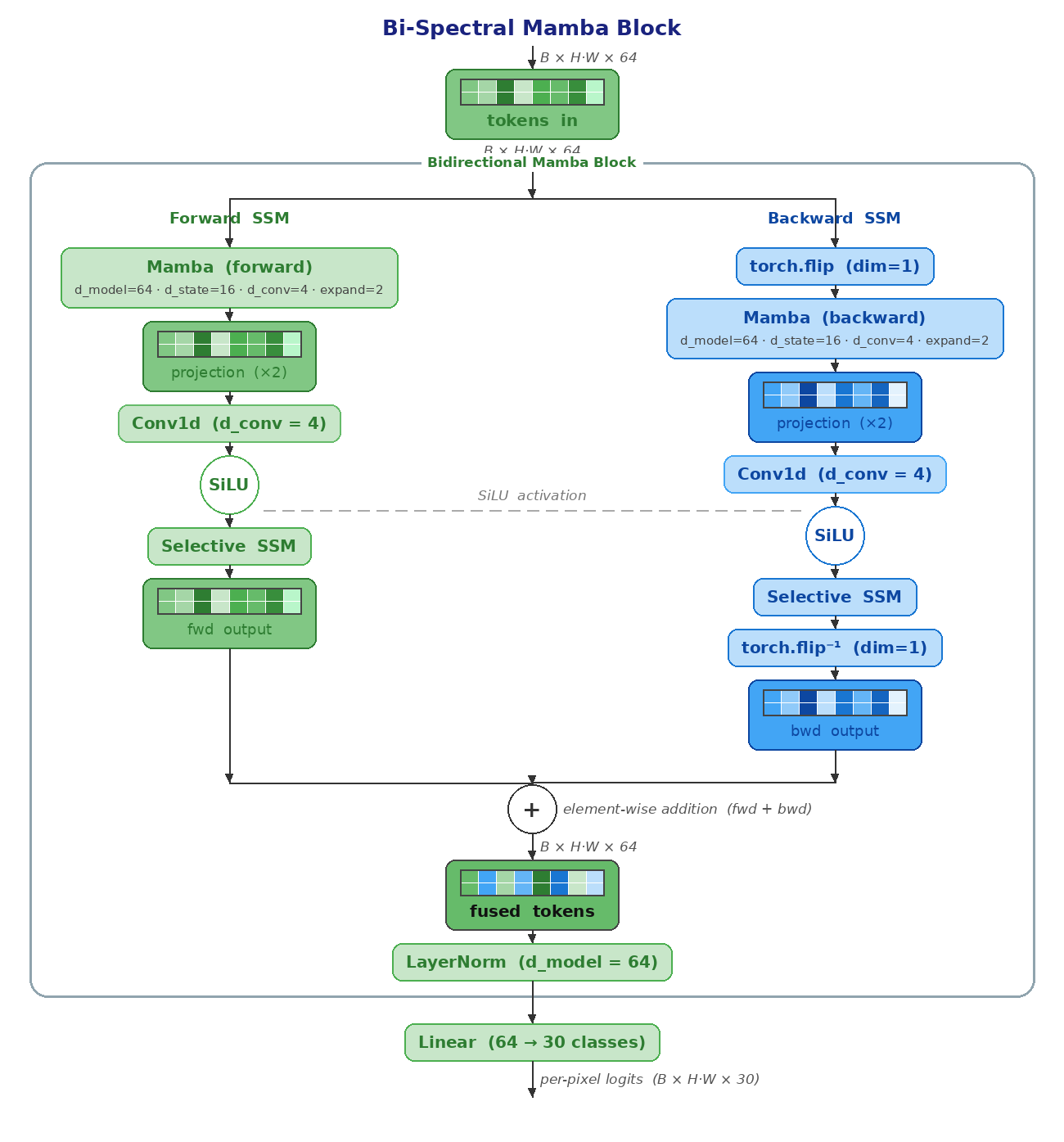}
    \caption{BiSpectral Mamba}
    \label{fig:bispecmambaarch}
\end{figure}

The architecture perfectly flattens the refined spatial tensor $F_{SE}$ into a continuous one-dimensional token sequence to enable high-speed state-space modeling as referenced in Fig~\ref{fig:bispecmambaarch}. This permutation reshapes the spatial axes into a sequence length $L = H \times W$, generating the tensor $T \in \mathbb{R}^{B \times L \times 64}$. Two independent Mamba blocks process this sequence to capture complex long-range spectral correlations. The forward block analyzes the sequence in its native spatial order. The backward block processes a mathematically inverted sequence. The system flips the backward output back to its original orientation and explicitly sums both representations to construct the fused tensor $H_{fused}$ .

\begin{equation}
    \vec{H} = \text{Mamba}_{fwd}(T)   
\end{equation}

\begin{equation}
    \overleftarrow{H} = \text{Flip}(\text{Mamba}_{bwd}(\text{Flip}(T)))    
\end{equation}

\begin{equation}
    H_{fused} = \vec{H} + \overleftarrow{H}    
\end{equation}

\subsection{Quantum Global Head Formulations}

The network explicitly splits the fused sequence into dual classification pathways. The quantum pathway begins by executing a global average pooling operation across the sequence dimension to extract a single holistic feature token $z_{global}$.

\begin{equation}
    z_{global} = \frac{1}{L} \sum_{i=1}^{L} H_{fused}^{(i)}
\end{equation}

\begin{figure}
    \centering
    \includegraphics[width=0.3\linewidth]{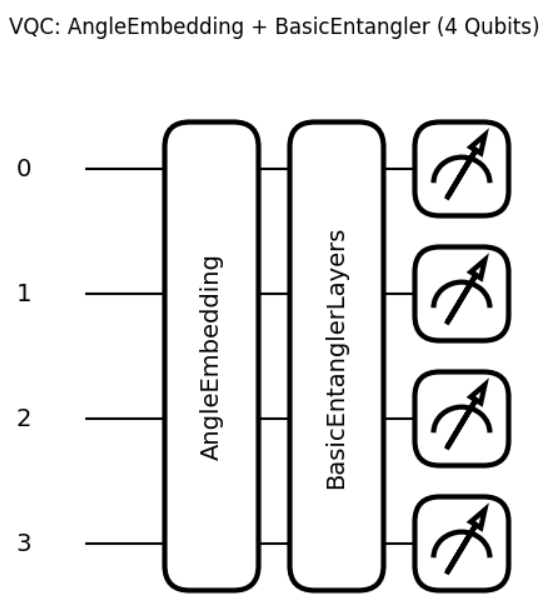}
    \caption{Quantum circuit diagram generated with help of the PennyLane Library}
    \label{fig:qcircuit}
\end{figure}

A CPU-pinned linear layer reduces this 64-dimensional token to exactly 4 dimensions, applying a hyperbolic tangent activation function to constrain the values.
\begin{equation}
    x_{red} = \tanh(W_{reduce} z_{global} + b_{reduce})
\end{equation}

The 4-qubit parameterized quantum circuit, as shown in Fig~\ref{fig:qcircuit} receives this dense vector. The circuit securely embeds the classical data via Angle Embedding across all four computational wires. Two successive Basic Entangler Layers process the embedded quantum states to model highly complex, non-linear global interactions.

\begin{equation}
    |\psi_{in}\rangle = \bigotimes_{i=1}^{4} R_x(x_{red}^{(i)}) |0\rangle
\end{equation}

\begin{equation}
    |\psi_{out}\rangle = U_{entangle}(\theta) |\psi_{in}\rangle
\end{equation}

$U_{entangle}(\theta)$ is a learnable unitary operator that processes compressed global features through rotation and entangling gates within a variational quantum circuit. It leverages superposition and entanglement to model complex, non-linear dependencies across the hyperspectral landscape that classical layers struggle to capture. This creates a quantum-enhanced global context that, when fused with local features, ensures high accuracy while maintaining parameter efficiency.

The circuit calculates the precise Pauli-Z expectation values for each independent qubit to extract the final quantum state representation $q$.
\begin{equation}
    q_i = \langle \psi_{out} | \sigma_z^{(i)}|\psi_{out} \rangle
\end{equation}

A final CPU-based linear layer formally projects these 4 quantum features into the target crop classes.
\begin{equation}
    Y_{quantum} = W_{q} q + b_{q}
\end{equation}

\textit{Hybrid Feature Fusion}:The classical pathway processes $H_{fused}$ through a LayerNorm and linear projection to generate a two-dimensional spatial classification map $Y_{classical}$ . The architecture expands the singular quantum logit vector $Y_{quantum}$ to perfectly match these spatial dimensions. A learnable parameter $\alpha$ dynamically regulates the structural contribution of the quantum features.
\begin{equation}
    Y_{logits} = Y_{classical} + \tanh(\alpha) Y_{quantum}
\end{equation}

The system executes a final bilinear interpolation to scale the logits flawlessly back to the original ground truth spatial dimensions

\section{Empirical Analysis}

\textit{Algorithm}: This is the experimental setup for the Quantum Enhanced Multi-Scale CNN with BiDirectional Mamba explained in the algorithm form as seen in Algorithm~\ref{alg:qcnn_bimamba}.

\begin{algorithm}[H]
\footnotesize
\caption{CNN-BiSpectralMamba-Quantum Execution}
\label{alg:qcnn_bimamba}
\begin{algorithmic}[1]
\REQUIRE Raw hyperspectral data cube $X \in \mathbb{R}^{B \times C \times H \times W}$, Target spatial maps $Y$

\STATE \textbf{Standardization:} Calculate per-band mean and standard deviation. Normalize $X$ using a strict $10^{-6}$ minimum variance floor

\STATE \textbf{Multi-Scale Extraction:} Process the normalized tensor through two CNN encoder blocks to generate spatial feature maps $E_1$ and $E_2$

\STATE \textbf{Decoder Fusion:} Bilinearly upsample $E_2$, concatenate with $E_1$, and apply a final convolution to yield the fused map $F_{dec}$

\STATE \textbf{Spectral Attention:} Apply the Spectral Squeeze-and-Excitation module to generate the recalibrated tensor $F_{SE}$

\STATE \textbf{Sequence Flattening:} Permute and flatten the spatial dimensions of $F_{SE}$ into a 1D token sequence $T$

\STATE \textbf{Bidirectional Mamba:} Process $T$ through a forward Mamba block and a flipped $T$ through a backward Mamba block.Un-flip the backward output and sum both representations to yield $H_{fused}$

\STATE \textbf{Classical Pathway:} Apply LayerNorm and a linear projection to $H_{fused}$ to generate the 2D classical spatial map $Y_{classical}$

\STATE \textbf{Quantum Pathway:} Compute the sequence mean to extract a global token, reduce to 4 dimensions, and execute the 4-qubit variational quantum circuit on the CPU to generate the global logit vector $Y_{quantum}$

\STATE \textbf{Fusion:} Expand $Y_{quantum}$ to match spatial dimensions and merge pathways using a learnable parameter $\alpha$:
\[
Y_{logits} = Y_{classical} + \tanh(\alpha)\, Y_{quantum}
\]

\STATE \textbf{Interpolation \& Optimization:} Bilinearly interpolate $Y_{logits}$ to the original ground truth dimensions. Compute the class-weighted Hybrid Cross-Entropy and Log-Cosh Dice Loss and update network weights

\end{algorithmic}
\end{algorithm}

\textit{HyperParameters}: The training engine strictly utilizes the AdamW optimizer to drive the network updates across exactly $100$ training epochs as seen in Table~\ref{tab:mamba_params}. The dataloader feeds the network using a fixed batch size of $8$ to maintain steady gradient flows without exceeding GPU memory constraints. The classical backbone and the quantum head operate on entirely decoupled learning rates to prevent massive classical gradients from destabilizing the sensitive quantum parameters. The classical multi-scale CNN and Mamba blocks update using a primary base learning rate of $3\times10^{-4}$. The quantum circuit processes its gradient updates at a drastically reduced learning rate of $5\times10^{-5}$. A universal weight decay penalty of $1\times10^{-4}$ permanently suppresses network overfitting across all training phases. The structural architecture strictly sets the core feature embedding dimension ($d_{model}$) to $64$ and utilizes a Squeeze-and-Excitation reduction ratio of $16$. The network initializes the dynamic quantum fusion parameter $\alpha$ exactly at $0.0$, allowing the model to learn the optimal quantum contribution dynamically through backpropagation. The hybrid loss function perfectly balances the weighted Cross-Entropy and Log-Cosh Dice metrics using a static weighting factor $\lambda$ of $0.5$. Finally, the data pipeline applies a $0.7$ power scaling to the inverse square root of the static class frequencies to generate the exact class weights required to penalize dominant vegetation types.

\begin{table}[t]
    \centering
    \caption{Training and architectural parameters for the CNN-BiSpectralMamba-Quantum model on UAV-HSI-Crop.}
    \footnotesize
    \label{tab:mamba_params}
    \begin{tabular}{lc}
    \toprule
    Parameter & Value \\
    \midrule
    Optimizer & AdamW \\
    Classical backbone learning rate & $3\times10^{-4}$ \\
    Quantum head learning rate & $5\times10^{-5}$ \\
    Epochs & $100$ \\
    Weight decay & $1\times10^{-4}$ \\
    Batch size & $8$ \\
    Feature embedding dimension ($d_{model}$) & $64$ \\
    SE block reduction ratio & $16$ \\
    Initial quantum weight parameter $\alpha$ & $0.0$ (Learnable) \\
    $\lambda$ & $0.5$ \\
    Class frequency power scaling & $0.7$ \\
    Normalization variance floor & $10^{-6}$ \\
    Dice loss smoothing factor $\epsilon$ & $10^{-5}$ \\
    \bottomrule
    \end{tabular}
\end{table}

\textit{Hardware Requirements}: This work was run on the Compute Canada NiBi cluster. This work provisioned the high-performance training environment using exactly $5$ cutting-edge NVIDIA H100 GPUs to aggressively accelerate the heavy multi-scale tensor convolutions and the Mamba sequence modeling. We were able to get a massive $1$ Terabyte of system memory alongside $25$ dedicated CPU cores to support the massive data pipelines. These robust, multi-threaded CPU resources directly handle the highly sensitive PennyLane quantum circuit calculations to permanently prevent GPU memory overflow during the complex state-vector simulations. The architecture successfully utilized this massive computing power to execute the complete convergence cycle within strict, highly efficient 8-hour continuous allocation windows.

\subsection{Loss Function}

We used a fusion of Class-Weighted Cross-Entropy and Log-Cosh Dice Loss proposed by Niu. et al.~\cite{cnn6}. This dual formulation perfectly balances pixel-level classification accuracy with the strict topological mapping of agricultural boundaries.

\textit{Class-Weighted Cross-Entropy ($L_{CE}$):}
The dataset inherently floods the network with dominant staple crops, masking the critical signatures of rare invasive weeds or diseased zones. I mathematically neutralized this severe data skew by actively calculating inverse-frequency class weights. The dataloader defines $N_c$ as the absolute pixel count for a specific class $c$. The algorithm applies a 0.7 power scaling to the inverse square root of these static counts, injecting a strict variance floor ($10^{-6}$) to permanently prevent zero-division errors during initialization:
\begin{equation}
W_c = \frac{1}{(\sqrt{N_c} + 10^{-6})^{0.7}}    
\end{equation}

The pipeline applies these aggressive penalty weights directly to the standard Cross-Entropy computation. For a true label $y$ and predicted probability $\hat{y}$, the network calculates the globally weighted pixel loss:
\begin{equation}
    L_{CE} = -\sum_{i=1}^{C} W_i y_i \log(\hat{y}_i)
\end{equation}

\textit{Log-Cosh Dice Loss ($L_{LCD}$):}
While Cross-Entropy optimizes individual pixels, it actively destroys the holistic geometry of overlapping vegetation rows. To permanently eliminate these boundary artifacts, the Dice coefficient was integrated to directly optimize the exact intersection-over-union metric. The formulation isolates the predictive overlap using distinct one-hot encoded target tensors $Y_{onehot}$ and prediction tensors $P$, strictly applying a smoothing factor $\epsilon = 10^{-5}$ to maintain structural stability:

\begin{equation}
    L_{DICE} = 1 - \frac{2 \sum (P \cdot Y_{onehot}) + \epsilon}{\sum P + \sum Y_{onehot} + \epsilon}
\end{equation}

Standard Dice loss notoriously generates highly aggressive, unstable gradients when evaluating tiny, minority crop patches. To force mathematical convergence, the system wraps the absolute Dice error in a logarithmic hyperbolic cosine operation. This non-linear transformation actively smooths the error surface, protecting the highly sensitive parameterized quantum circuit from gradient shock:
\begin{equation}
    L_{LCD} = \log(\cosh(L_{DICE}))
\end{equation}

\textit{Total Hybrid Fusion:}
The optimization engine cleanly fuses these two distinct mathematical penalties into a single, cohesive gradient update. The pipeline utilizes a static weighting parameter $\lambda$ set exactly to $0.5$ to balance the classical pixel accuracy with the structural boundary mapping equally:

\begin{equation}
L_{total} = (1 - \lambda) L_{CE} + \lambda L_{LCD}    
\end{equation}

\subsection{Evaluation Metrics}

\textit{Overall Accuracy (OA):}
Overall Accuracy establishes the fundamental baseline of the model's global predictive power~\cite{cnn6}. The metric mathematically divides the exact number of correctly classified pixels by the absolute total pixel population across the entire spatial grid. We define $C$ as the total number of classes (30), $n_{ii}$ as the exact count of true-positive pixels for a specific class, and $N$ as the total spatial pixel count:

\begin{equation}
    OA = \frac{\sum_{i=1}^{C} n_{ii}}{N}
\end{equation}

\textit{Cohen's Kappa Coefficient ($\kappa$):}
Because the UAV-HSI-Crop dataset possesses massive class imbalances, a model could simply guess the dominant crop class and achieve a deceivingly high Overall Accuracy~\cite{cnn6}. Cohen's Kappa ruthlessly mathematically purifies the performance score by factoring out any expected agreement caused by pure random chance. This provides an unbreakable statistical reliability score for the entire framework.The formula requires the observed agreement $p_o$ (which directly equals the calculated OA) and the statistically expected chance agreement 
\begin{equation}
p_e:\kappa = \frac{p_o - p_e}{1 - p_e}   
\end{equation}

To strictly calculate $p_e$, the algorithm explicitly cross-multiplies the total ground truth pixels $n_{g,i}$ with the total predicted pixels $n_{p,i}$ for every single class $i$, completely neutralizing the statistical skew of dominant vegetation types before dividing by the squared total pixel count:

\begin{equation}
p_e = \frac{\sum_{i=1}^{C}(n_{g,i} \cdot n_{p,i})}{N^2}   
\end{equation}

~~~~~The empirical framework detailed above guarantees that our recorded performance gains are structurally legitimate, rather than statistical illusions generated by extreme dataset skew. By mathematically forcing the networks to respect and map minority crop boundaries via the Log-Cosh Dice loss, and by rigorously purifying the final accuracy scores through Cohen's Kappa, we eliminated the deceptive metric inflation that plagues standard agricultural deep learning research and with these experimental baseline firmly established, the subsequent results definitively prove exactly how these lightweight hybrid architectures can match and overtake traditional, parameter-heavy classical models while being more efficient.

\section{Experimental Results}

~~~~~In this section, we will go over the results of the experimental setup and compare our model with different models that followed similar loss and evaluation metrics, and see how our models stack up in comparison with them.

\subsection{Comparison with Existing Models}
\begin{figure}
    \centering
    \includegraphics[width=0.6\linewidth]{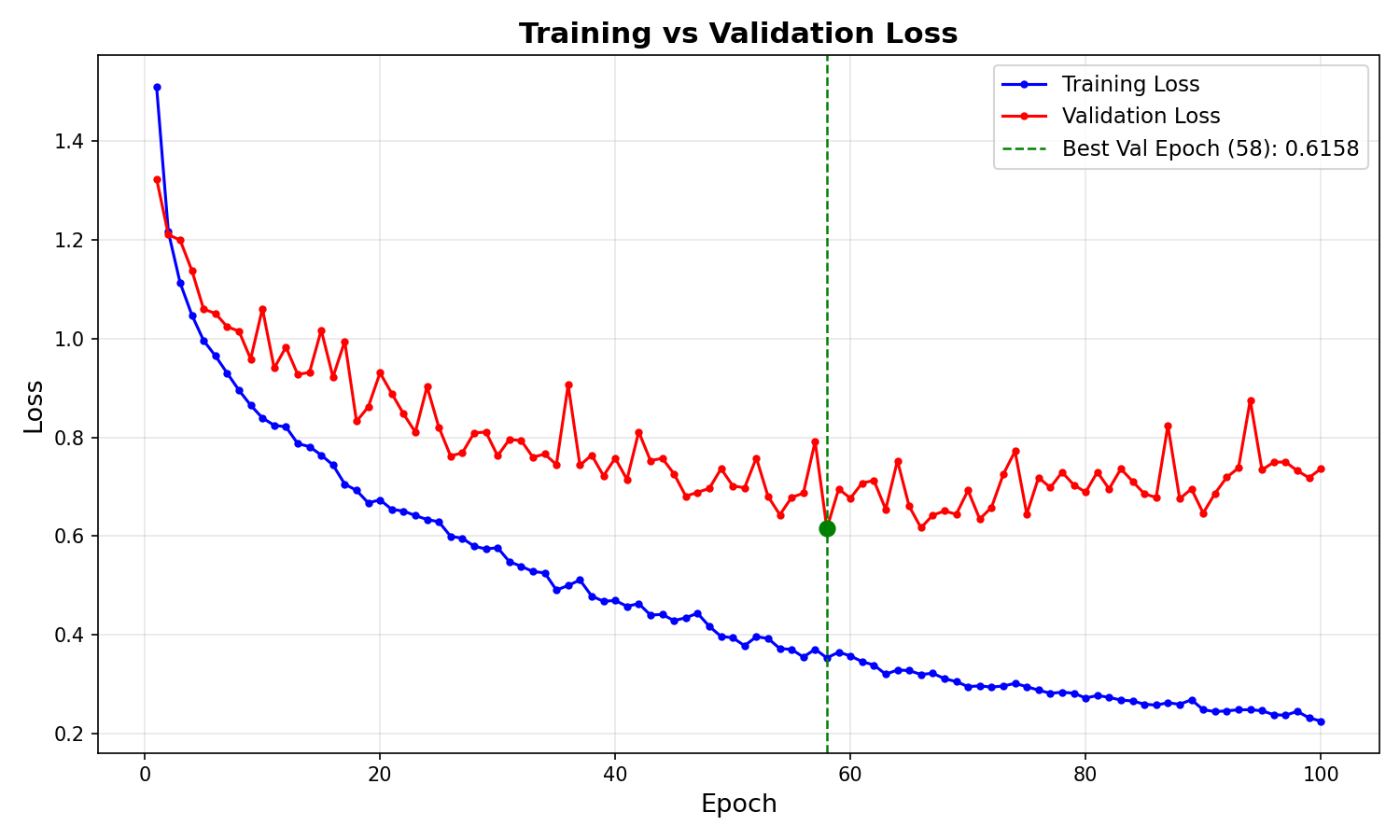}
    \caption{Quantum Enhanced Multi-Scale CNN with BiDirectional Mamba Training Loss}
    \label{fig:Main Training Loss}
\end{figure}

In this section, we compare the proposed CNN-BiSpectralMamba-Quantum architecture against established UAV-HSI-Crop baselines, incorporating the results from our preceding QPGF framework for direct evolutionary comparison. Table ~\ref{tab:comparison_mamba} provides the numbers behind this comparison. The loss function for the main training is referenced in Fig~\ref{fig:Main Training Loss}.

\begin{table*}[htbp]
    \centering
    \caption{Performance comparison on UAV-HSI-Crop.}
    \footnotesize
    \label{tab:comparison_mamba}
    \begin{tabular}{lcccc}
    \toprule
    Model & Epochs &Parameters(Millions) & OA (\%)$\uparrow$ & Kappa ($\kappa \times 100$)$\uparrow$ \\
        \midrule
    SegNet~\cite{cnn6} & $200$ & $29M$& $43.61$ & $54.15$ \\
    SETR~\cite{cnn6}  & $200$ &$43M$&$69.47$ & $72.67$ \\
    UNet~\cite{cnn6}  & $200$ &$8M$&$76.07$ & $71.31$ \\
    TransUNet~\cite{cnn6}  & $200$&$105M$& $78.64$ & $74.56$ \\
    HSI-TransUNet~\cite{cnn6} & $200$ &$110M$& $86.05$ & $83.47$\\
    HRS-UNet~\cite{cnn5} & $200$ & Not Reported & $89.96$ & $88.14$ \\
    QPGF (ours) & 200 &4.1M&81.92&78.41 \\
    \textbf{Quantum Enhanced CNN with Mamba(Ours)}  &\textbf{200}&\textbf{0.24M} & \textbf{84.83} & \textbf{82.07} \\
    \bottomrule
    \end{tabular}
\end{table*}

The results demonstrate a massive evolutionary leap from our initial graph-based approach. By utilizing an optimal quantum fusion parameter of $\alpha=0.2$, the CNN-BiSpectralMamba-Quantum model decisively outperforms standard transformer architectures like TransUNet and eclipses our own QPGF baseline by nearly $3\%$. Although it does not mathematically exceed the peak OA of the massive HSI-TransUNet or HRS-UNet models, it reaches a highly competitive threshold in way lesser parameters of $246,793$. The comparison highlights an important trade-off: we actively bypassed the unsustainable parameter bloat inherent to those purely classical networks, opting instead for a highly optimized operating point. The model delivers aggressive sequence modeling for high spatial-spectral fidelity, perfectly stabilized by a lightweight quantum global head.

\subsection{Ablation Study}

\begin{figure}[t]
    \centering
    \subfigure[Ablation-1]{\includegraphics[width=0.45\linewidth]{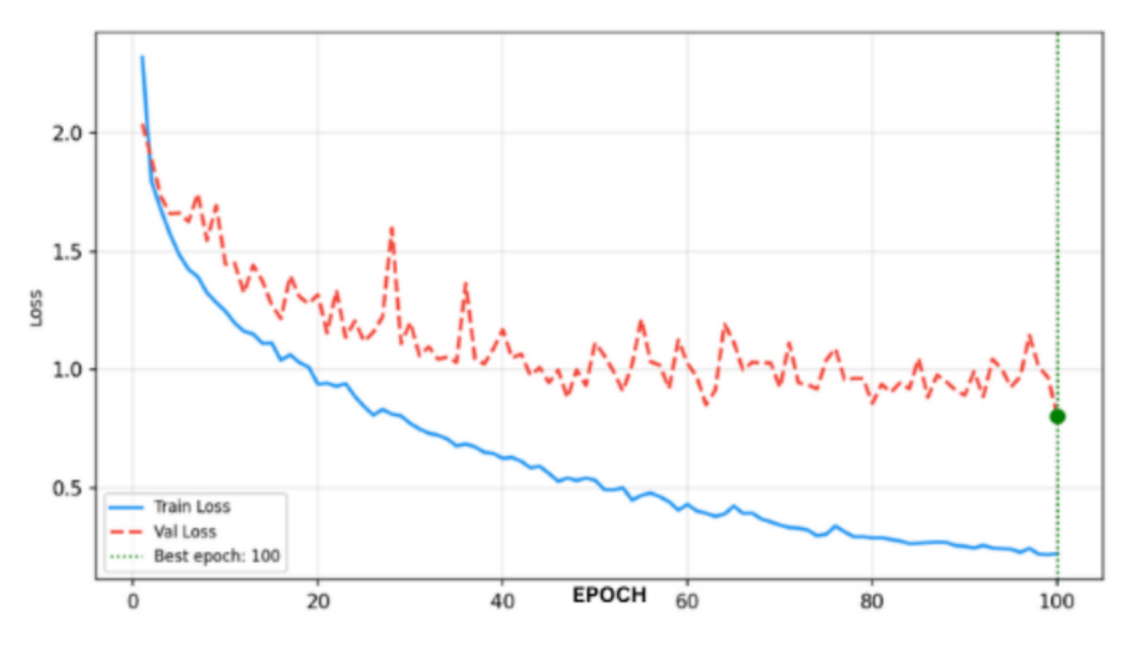}}
    \hfill
    \subfigure[Ablation-2]{\includegraphics[width=0.45\linewidth]{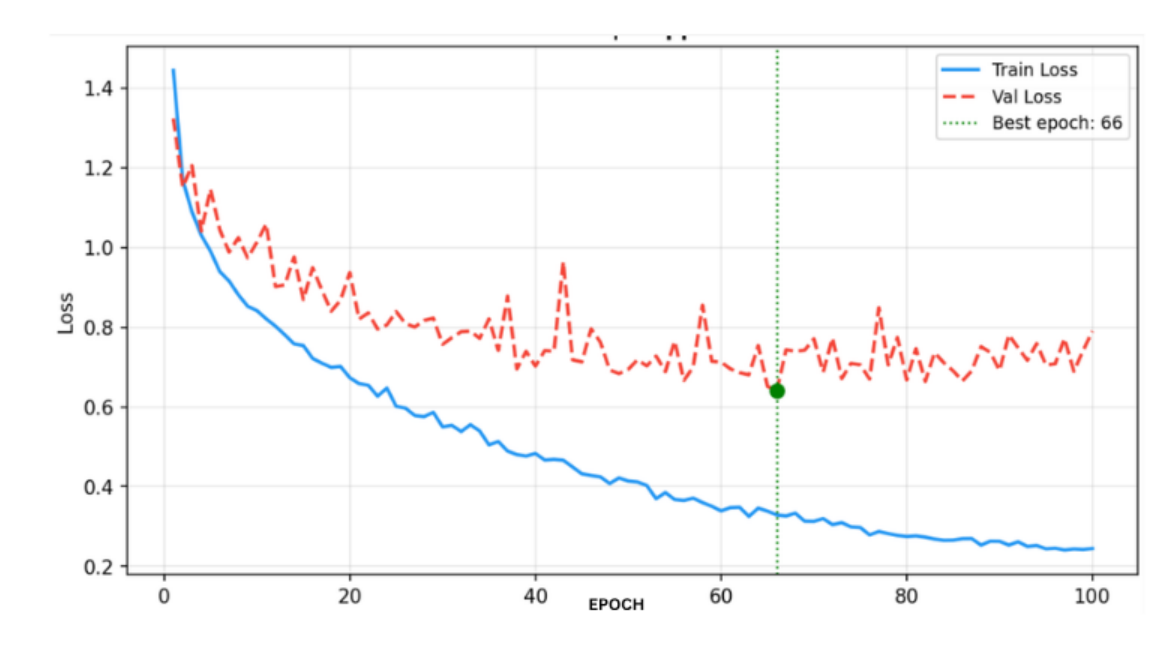}}
    \subfigure[Ablation-3]{\includegraphics[width=0.45\linewidth]{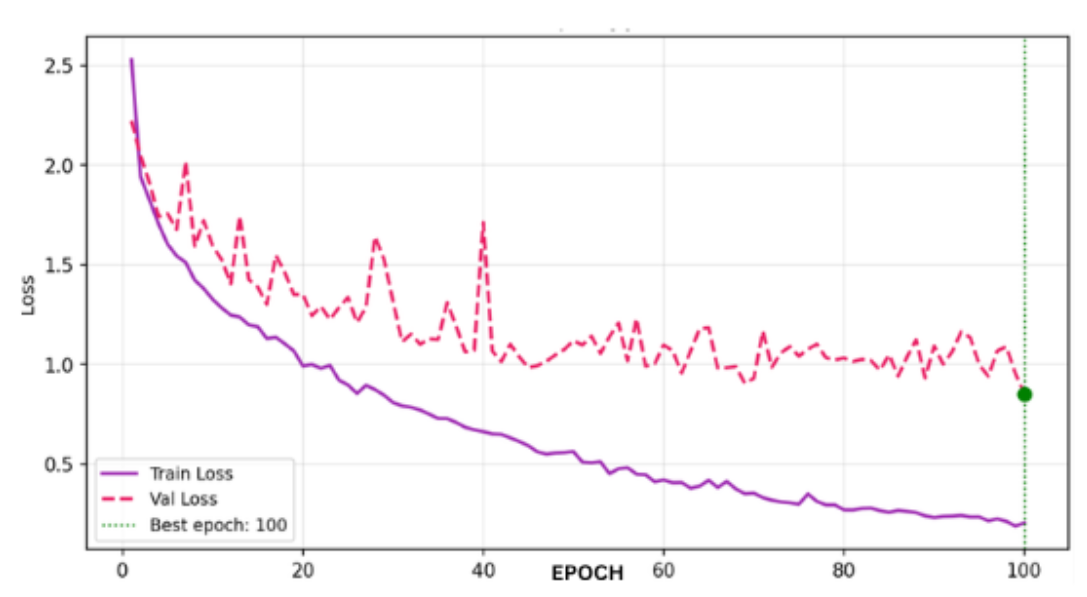}}
    \hfill
    \subfigure[Ablation-4]{\includegraphics[width=0.45\linewidth]{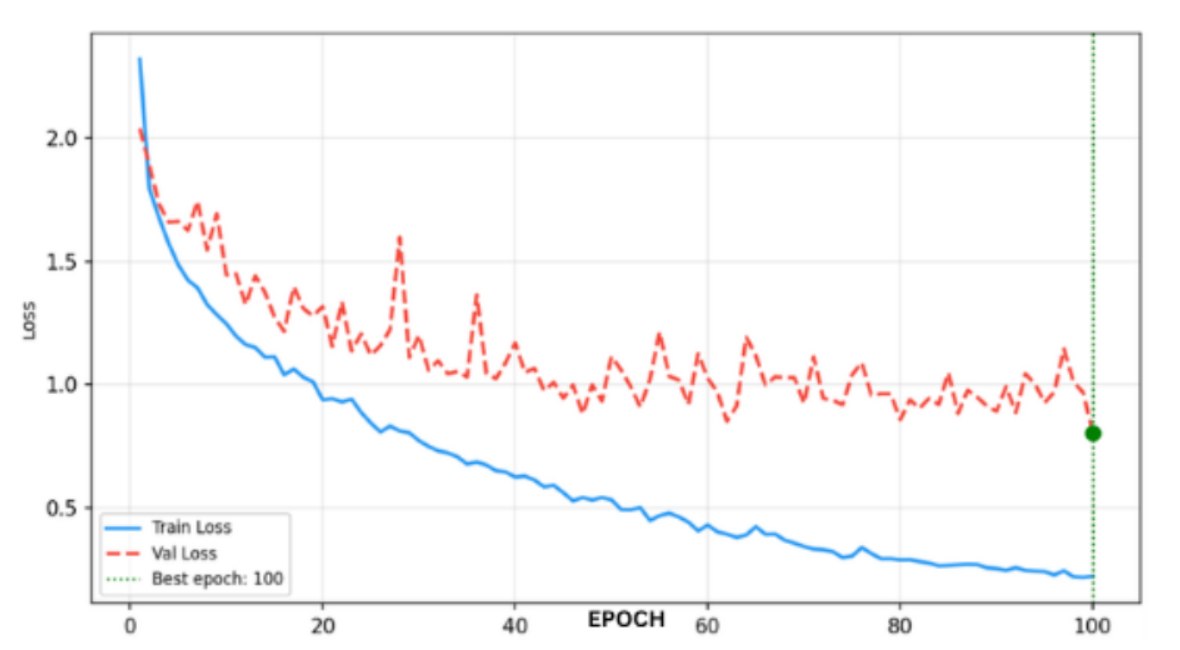}}
    \caption{Training loss curves across all ablation variants.}
    \label{fig:training_losses_mambaablation}
\end{figure}

To rigorously validate the individual architectural components, we executed four targeted ablation settings against our final proposed model. \textit{Ablation 1 (No Quantum)} removes the 4-qubit variational quantum circuit. \textit{Ablation 2 (No BiMamba)} removes the bidirectional state-space sequence modeling blocks entirely. \textit{Ablation 3 (Plain CE)} disables the hybrid Log-Cosh Dice boundary loss, relying strictly on standard Cross-Entropy. \textit{Ablation 4 (No SE)} removes the Spectral Squeeze-and-Excitation channel recalibration module.Training losses are referenced in Fig~\ref{fig:training_losses_mambaablation}.

\begin{table}[htbp]
    \centering
    \caption{Ablation results for CNN-BiSpectralMamba-Quantum on UAV-HSI-Crop.}
    \footnotesize
    \label{tab:mamba_ablation}
    \begin{tabular}{lcc}
    \toprule
    Model Variant & OA (\%)$\uparrow$ & Kappa ($\kappa \times 100$)$\uparrow$ \\
    \midrule
    No Quantum (Ablation 1) & $83.51$ & $80.65$ \\
    No BiMamba (Ablation 2) & $79.50$ & $75.59$ \\
    Plain CE (Ablation 3) & $85.24$ & $82.64$ \\
    No SE (Ablation 4) & $84.37$ & $81.56$ \\
    \bottomrule
    \end{tabular}
\end{table}

The ablation trend in Table ~\ref{tab:mamba_ablation} provides a highly illuminating structural interpretation. The catastrophic drop observed in Ablation 2 ($-5.33\% OA$) definitively proves that bidirectional Mamba sequence modeling forms the absolute critical core of the architecture's feature extraction capability. Reintroducing the quantum global head (Proposed vs. Ablation 1) actively pushes the OA from $83.51\%$to $84.83\%$, mathematically validating our hypothesis that the parameterized entanglement circuit successfully captures holistic field geometry that the classical backbone misses. The Squeeze-and-Excitation block (Ablation 4) provides a smaller structural refinement ($+0.46\% OA$).

The unique data point within the ablation study is Ablation 3 (Plain CE), which achieves a numerically higher global Overall Accuracy ($85.24\%$). Rather than a failure of the proposed loss function, this highlights a critical behavioral trade-off inherent to severely imbalanced agricultural datasets. Plain Cross-Entropy easily inflates the global pixel-counting metric by aggressively optimizing for massive, dominant crop classes, often sacrificing the precise topological mapping of rare vegetation. Our Proposed model ($84.83\%$) enforces the Hybrid Log-Cosh Dice loss to actively penalize dominant classes and strictly enforce boundary intersection-over-union. We sacrifice a marginal $0.41\%$ of skewed global accuracy to guarantee a highly equitable, structurally robust mapping of the entire minority crop landscape.

\subsection{Error Analysis}

\begin{figure}
    \centering
    \includegraphics[width=0.98\linewidth]{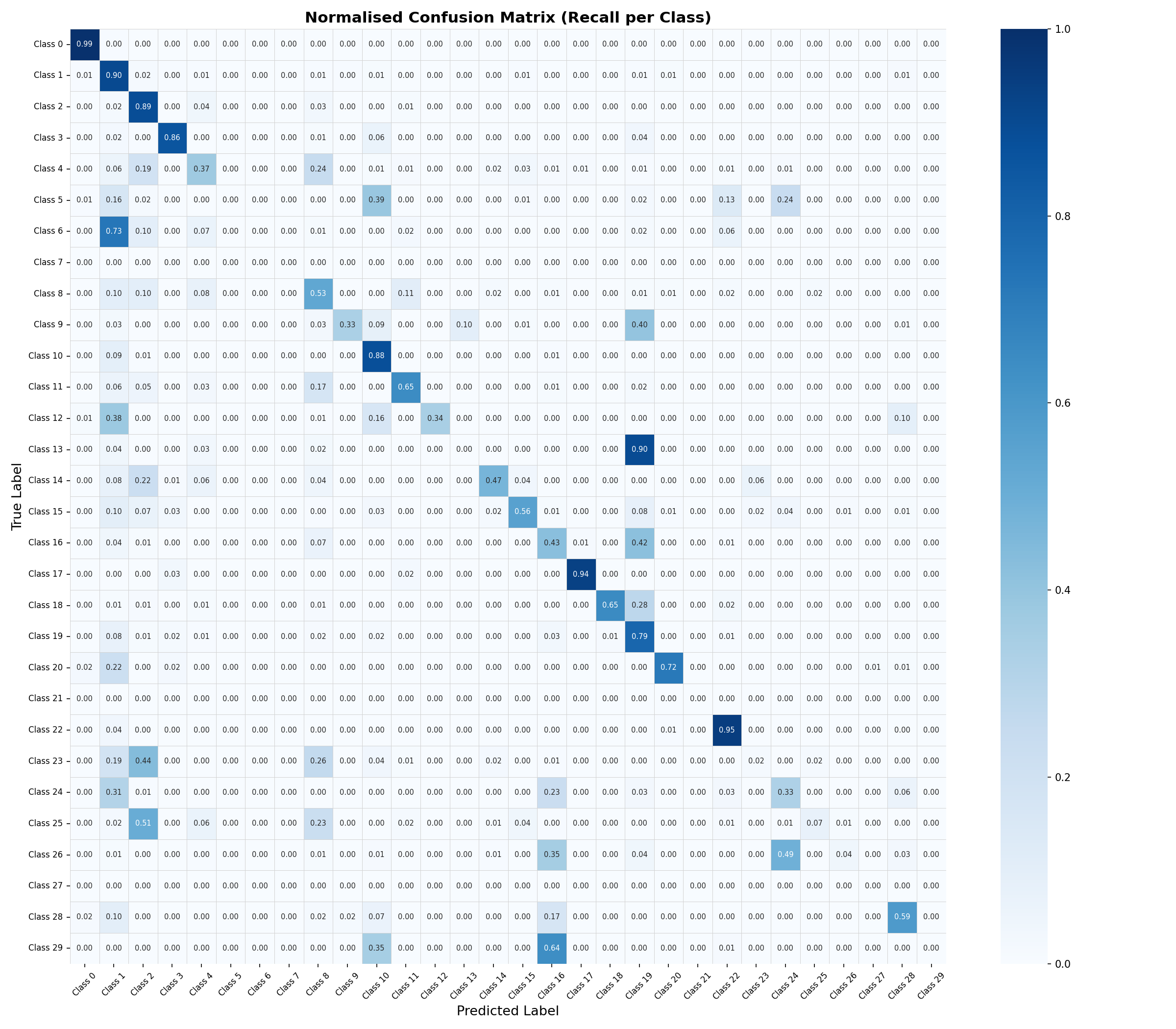}
    \caption{Confusion Matrix For Quantum Enhanced CNN with Mamba }
    \label{fig:mambaconfusion}
\end{figure}

Similar to the QPGF framework, the model performs exceptionally well on major crop categories while exhibiting expected sensitivities at complex boundary regions. Because the bidirectional Mamba sequence aggressively tracks contiguous spectral signatures as seen in the Fig~\ref{fig:mambaconfusion}, slight misclassifications still occur where phenologically similar crops drastically overlap in physical space. However, the integration of the multi-scale CNN successfully prevents the Mamba blocks from completely destroying local spatial geometry, a massive improvement over pure state-space models.

While the $84.83\%$ OA proves the architectural fusion is highly successful, future gains will likely emerge from increasing the quantum qubit count and utilizing dynamic, class-aware spatial priors during the final bilinear upsampling phase.

\section{Conclusion}

In this work, we outlined a BiSpectral Mamba-based architecture for HSI crop analysis. This framework begins with a multi-scale CNN backbone that extracts spatial-spectral features using a hierarchy of convolutional layers and feature fusion across resolutions.
A spectral channel attention mechanism refines these features by emphasizing bands and suppressing redundant or noisy channels. The refined features are then fed into a Mamba module that treats spatial-spectral feature maps as sequences of tokens and models long-range dependencies in both directions along the sequence. We can also conclude from this work that.
First HSI crop analysis benefits from architectures that explicitly structure the data, such as state-space formulations, rather than treating hyperspectral cubes as flat images or generic sequences. Second, combining approaches at multiple levels appears to be more effective and practical than trying to replace entire pipelines with a single paradigm.
Third, optimization and loss design in the presence of class imbalance and limited labels can have effects on performance comparable to those of architectural modifications. The combination of loss-class-weighted strategies and carefully tuned fusion coefficients played a central role in stabilizing training and improving final metrics.
Finally, while quantum and quantum-inspired components are still being explored, their integration as global-context modules in otherwise classical models is a promising way to test their utility without incurring high costs. The use of BiSpectral Mamba architecture for HSI crop analysis shows that this approach can be effective and efficient.
The proposed framework can be used for agricultural applications, including crop disease analysis, crop yield prediction, and soil moisture estimation. The architecture can be further improved by incorporating advanced graph-based models, state-space formulations, and quantum-inspired components.
The results of this work can be used to develop accurate and efficient models for HSI crop analysis, which can help farmers and agricultural professionals make better decisions and improve crop yields. The use of architectures and quantum-inspired components can also be applied to other areas of remote sensing and computer vision.
Overall, this work demonstrates the potential of deep learning architecture for HSI crop analysis and highlights the importance of structuring the data by combining different approaches and optimizing the models for better performance. The proposed frameworks can be used as a starting point for research and development in this area.

\bibliographystyle{unsrt}
\bibliography{references_}

@misc{quantumefficiency,
      title={Let the Quantum Creep In: Designing Quantum Neural Network Models by Gradually Swapping Out Classical Components}, 
      author = {Peiyong Wang and Casey. R. Myers and Lloyd C. L. Hollenberg and Udaya Parampalli},
      year={2024},
      eprint={2409.17583},
      archivePrefix={arXiv},
      primaryClass={quant-ph},
      url={https://arxiv.org/abs/2409.17583}, 
}

@article{multiscalecnn,
    author = {Li S and Zhu X and Bao J},
    title = {Hierarchical Multi-Scale Convolutional Neural Networks for Hyperspectral Image Classification},
    journal = {Sensors (Basel)},
    year = {2019},
    url={ https://doi.org/10.3390/s19071714}
}

@article{bidimamba,
    author = {Sun M and Zhang J and He X and Zhong Y},
    title = {Bidirectional Mamba with Dual-Branch Feature Extraction for Hyperspectral Image Classification},
    journal = { Sensors (Basel).},
    year = {2024},
    url={https://doi.org/10.3390/s24216899}
}

@article{cnn1,
    author = {Guo, X. and Feng, Q. and Guo, F.},
    title = {CMTNet: a hybrid CNN-transformer network for UAV-based hyperspectral crop classification in precision agriculture.},
    journal = {Sci Rep 15, 12383 },
    year = {2025},
    url={https://doi.org/10.1038/s41598-025-97052-w}
}

@article{cnn2,
    author = {R.Ablin and G.Prabin} ,
    title = {An optimal model using hybrid LCNN- GRU for efficient hyperspectral image classification},
    publisher = {Springer},
    journal={International Journal of Information Technology},
    volume={15},
    year = {2023},
    url={https://link.springer.com/article/10.1007/s41870-023-01317-4}}

@inproceedings{cnn5,
    author = {Zhiyu Yang and Lei Zou and Yuhuai Lin},
    title = {HRS-UNet: A Semantic Segmentation Model for Precise Crop Classification in Hyperspectral Remote Sensing Image},
    booktitle = {Proceedings of the 21st International Conference on Intelligent Computing (ICIC 2025)},
    month = {July},
    date = {26-29},
    year = {2025},
    address = {Ningbo, China},
    pages = {2129-2140},
    note = {Poster Volume II},
    doi = {10.65286/icic.v21i2.73126},
    url={http://poster-openaccess.com/files/ICIC2025/4227.pdf}
}

@article{cnn6,
	author = {Bowen Niu and Quanlong Feng and Boan Chen and Cong Ou and Yiming Liu and Jianyu Yang},
	journal = {Computers and Electronics in Agriculture},
	year = {2022},
	month = {October},
	url ={https://www.sciencedirect.com/science/article/pii/S0168169922006093},
	publisher = {https://www.sciencedirect.com/science/article/pii/S0168169922006093},
	title = {HSI-TransUNet: A transformer based semantic segmentation model for crop mapping from UAV hyperspectral imagery},
}

@misc{cnn7,
      title={Explaining hyperspectral imaging based plant disease identification: 3D CNN and saliency maps}, 
      author = {Koushik Nagasubramanian and Sarah Jones and Asheesh K. Singh and Arti Singh and Baskar Ganapathysubramanian and Soumik Sarkar},
      year={2018},
      eprint={1804.08831},
      archivePrefix={arXiv},
      primaryClass={cs.CV},
      url={https://arxiv.org/abs/1804.08831}, 
}

@article{cnn9,
    author = {S P and Shirly Edward A},
    title = {MLVI-CNN: a hyperspectral stress detection framework using machine learning-optimized indices and deep learning for precision agriculture},
    journal = {Front. Plant Sci},
    year = {2025},
    doi={10.3389/fpls.2025.1631928},
    url={https://www.frontiersin.org/journals/plant-science/articles/10.3389/fpls.2025.1631928/full}
}

@inproceedings{trans2,
author = {Mazharul Hossain and Aaron Robinson and Lan Wang and Chrysanthe Preza},
title = {{Improving semantic segmentation through task adaptation for UAV hyperspectral agricultural imagery}},
volume = {13475},
booktitle = {Autonomous Air and Ground Sensing Systems for Agricultural Optimization and Phenotyping X},
editor = {J. Alex Thomasson and Christoph Bauer},
organization = {International Society for Optics and Photonics},
publisher = {SPIE},
pages = {1347507},
year = {2025},
doi = {10.1117/12.3053426},
URL = {https://doi.org/10.1117/12.3053426}
}

@article{trans3,
    author = {Pi W and Zhang T and Wang R and Ma G and Wang Y and Du J} ,
    title = {Semantic-Guided Transformer Network for Crop Classification in Hyperspectral Images},
    journal = { J Imaging.},
    year = {2025},
    doi={10.3390/jimaging11020037},
    url={https://doi.org/10.3390/jimaging11020037}
}

@article{trans4,
    author = {Xie J and Hua J and Chen S and Wu P and Gao P and Sun D and Lyu Z and Lyu S and Xue X and Lu J},
    title = {HyperSFormer: A Transformer-Based End-to-End Hyperspectral Image Classification Method for Crop Classification},
    journal = {Remote Sensing},
    year = {2023},
    url={https://doi.org/10.3390/rs15143491}
}

@article{trans5,
    author = {Gu Q and Luan H and Huang K and Sun Y},
    title = {Hyperspectral Image Classification Using Multi-Scale Lightweight Transformer.},
    journal = {Electronics},
    year = {2024},
    url={https://doi.org/10.3390/electronics13050949}
}

@article{trasn6,
    author = {Linsheng Huang and Lu Zhang and Chao Ruan and Jinling Zhao},
    title = {SSSAT-Net: Spectral-Spatial Self-Attention-Based Transformer Network },
    journal = {Optics and Lasers in Engineering},
    year = {2025},
    doi={https://doi.org/10.1016/j.optlaseng.2025.109154},
    url={https://www.sciencedirect.com/science/article/pii/S0143816625003392}    
}

@ARTICLE{trans7,
  author = {Huang, Shaoguang and Xiao, Wei and Chen, Hongyu and Bejo, Siti Khairunniza and Zhang, Hongyan},
  journal={IEEE Transactions on Geoscience and Remote Sensing}, 
  title={Hyperspectral Image Classification Based on a Locally Enhanced Transformer Network}, 
  year={2025},
  volume={63},
  number={},
  pages={1-17},
  doi={10.1109/TGRS.2025.3566672},
  url={https://ieeexplore.ieee.org/document/10985782}}

@ARTICLE{trans8,
  author = {Wang, Yuyang and Shu, Zhenqiu and Yu, Zhengtao},
  journal={IEEE Journal of Selected Topics in Applied Earth Observations and Remote Sensing}, 
  title={Efficient Attention Transformer Network With Self-Similarity Feature Enhancement for Hyperspectral Image Classification}, 
  year={2025},
  volume={18},
  number={},
  pages={11469-11486},
  doi={10.1109/JSTARS.2025.3560384},
  url={https://ieeexplore.ieee.org/document/10964176}}

@article{trans9,
    author = {Liu, Y. and Dian, R. and Li, S.},
    title = { Low-Rank Transformer for High-Resolution Hyperspectral Computational Imaging},
    journal = {Int J Comput Vis 133},
    year = {2025},
    url={https://doi.org/10.1007/s11263-024-02203-7}
}

@article{trans10,
    author = {Alireza Sanaeifar and Shahryar Kianian and Ruth Dill-Macky and Susan Reynolds and Matthew J Moscou and Rebecca D. Curland and James Anderson and Matthew N. Rouse and Ce Yang},
    title = {Transformer-based and band-selected models for UAV hyperspectral wheat disease classification},
    journal = {Smart Agricultural Technology},
    year = {2026},
    doi={https://doi.org/10.1016/j.atech.2025.101714.},
    url={https://www.sciencedirect.com/science/article/pii/S2772375525009451)}
}

@article{quant1,
    author = {Abhishek Chandrakant Nikam and Rahul Borate and Masira Kulkarni and Aayush Patil and Sadaf Shaikh},
    title = {Fruit Crop Disease Classification Using Quantum Machine Learning: A Pilot Study},
    journal = {Quantum Journal of Engineering, Science and Technology,},
    year = {2025},
    url={https://qjoest.com/index.php/qjoest/article/view/257}
}

@ARTICLE{quant2,
  author = {Otgonbaatar, Soronzonbold and Datcu, Mihai},
  journal={IEEE Journal of Selected Topics in Applied Earth Observations and Remote Sensing}, 
  title={A Quantum Annealer for Subset Feature Selection and the Classification of Hyperspectral Images}, 
  year={2021},
  volume={14},
  number={},
  pages={7057-7065},
  doi={10.1109/JSTARS.2021.3095377},
  url={https://ieeexplore.ieee.org/document/9477115}}

@article{quant3,
   title={Underdetermined Blind Source Separation via Weighted Simplex Shrinkage Regularization and Quantum Deep Image Prior},
   volume={35},
   ISSN={1941-0042},
   url={http://dx.doi.org/10.1109/TIP.2026.3673957},
   DOI={10.1109/tip.2026.3673957},
   journal={IEEE Transactions on Image Processing},
   publisher={Institute of Electrical and Electronics Engineers (IEEE)},
   author = {Lin, Chia-Hsiang and Young, Si-Sheng},
   year={2026},
   pages={3069--3084} }

@article{quant4,
    author = {Jie Zhang and Yongshan Zhang and Yicong Zhou1},
    title = {Quantum-Inspired Spectral-Spatial Pyramid Network for Hyperspectral ImageClassification},
    journal ={IEEE} ,
    year = {2023},
    url={https://ieeexplore.ieee.org/document/10203069}
}

@article{quant5,
  author = {Yi Liu and Wendy Wang and Haibo Wang and Bahram Alidaee},
  title = {Quantum Machine Learning on Remote Sensing Data Classification},
  journal = {Journal of Engineering Research and Sciences},
  year = {2023},
  volume = {2},
  number = {12},
  pages = {23--33},
  doi = {10.55708/js0212004},
  url={https://www.jenrs.com/v02/i12/p004/}
}

@misc{quant6,
      title={HyperKING: Quantum-Classical Generative Adversarial Networks for Hyperspectral Image Restoration}, 
      author = {Chia-Hsiang Lin and Si-Sheng Young},
      year={2025},
      eprint={2504.11782},
      archivePrefix={arXiv},
      primaryClass={eess.IV},
      url={https://arxiv.org/abs/2504.11782}, 
}

@article{mamba1,
    author = {Huang L and Chen Y and He X} ,
    title = {Spectral-Spatial Mamba for Hyperspectral Image Classification},
    journal = {Remote Sensing.},
    year = {2024},
    url={https://doi.org/10.3390/rs16132449}
}

@article{mamba2,
    author = {Liao J and Wang L},
    title = {HyperspectralMamba: A Novel State Space Model Architecture for Hyperspectral Image Classification},
    journal = {Remote Sensing. },
    year = {2025},
    url={https://doi.org/10.3390/rs17152577}
}

@article{mamba4,
    author = {Zhang Y and Jin X and Zhang X and Wu Y and Tu L},
    title = {EchoMamba: A new Mamba model for fast and efficient hyperspectral image classification.EchoMamba: A new Mamba model for fast and efficient hyperspectral image classification.},
    journal = {PLoS One},
    year = {2025},
    url={https://journals.plos.org/plosone/article?id=10.1371/journal.pone.0330678}
}

@article{mamba5,
   title={Spatial--spectral morphological mamba for hyperspectral image classification},
   volume={636},
   ISSN={0925-2312},
   url={http://dx.doi.org/10.1016/j.neucom.2025.129995},
   DOI={10.1016/j.neucom.2025.129995},
   journal={Neurocomputing},
   publisher={Elsevier BV},
   author = {Ahmad, Muhammad and Butt, Muhammad Hassaan Farooq and Khan, Adil Mehmood and Mazzara, Manuel and Distefano, Salvatore and Usama, Muhammad and Roy, Swalpa Kumar and Chanussot, Jocelyn and Hong, Danfeng},
   year={2025},
   month=jul, pages={129995} }

@article{mamba6,
   title={S2Mamba: A Spatial--Spectral State Space Model for Hyperspectral Image Classification},
   volume={63},
   ISSN={1558-0644},
   url={http://dx.doi.org/10.1109/TGRS.2025.3530993},
   DOI={10.1109/tgrs.2025.3530993},
   journal={IEEE Transactions on Geoscience and Remote Sensing},
   publisher={Institute of Electrical and Electronics Engineers (IEEE)},
   author = {Wang, Guanchun and Zhang, Xiangrong and Peng, Zelin and Zhang, Tianyang and Jiao, Licheng},
   year={2025},
   pages={1--13} }

@misc{mamba7,
      title={MambaMoE: Mixture-of-Spectral-Spatial-Experts State Space Model for Hyperspectral Image Classification}, 
      author = {Yichu Xu and Di Wang and Hongzan Jiao and Lefei Zhang and Liangpei Zhang},
      year={2025},
      eprint={2504.20509},
      archivePrefix={arXiv},
      primaryClass={cs.CV},
      url={https://arxiv.org/abs/2504.20509}, 
}

@misc{mamba8,
      title={SSUMamba: Spatial-Spectral Selective State Space Model for Hyperspectral Image Denoising}, 
      author = {Guanyiman Fu and Fengchao Xiong and Jianfeng Lu and Jun Zhou},
      year={2024},
      eprint={2405.01726},
      archivePrefix={arXiv},
      primaryClass={eess.IV},
      url={https://arxiv.org/abs/2405.01726}, 
}

@misc{vanillamambba,
  author = {Anish Bhardwaj},
  title  = {What is a Mamba model},
  url    = {https://www.geeksforgeeks.org/artificial-intelligence/what-is-a-mamba-model/},
  note = {Available at: \url{https://www.geeksforgeeks.org/artificial-intelligence/what-is-a-mamba-model/} (accessed 2026-02-01)}
}

\end{document}